\documentclass[letterpaper]{article}

\usepackage[hidelinks]{hyperref}
\usepackage{orcidlink}
\usepackage{amsmath}
\usepackage{amssymb}
\usepackage{amsfonts}
\usepackage{newfloat}
\usepackage{listings}
\usepackage{xcolor}
\usepackage{siunitx}
\usepackage{todonotes}
\usepackage{array}
\usepackage{subcaption}
\usepackage{booktabs}
\usepackage{microtype}
\usepackage{tabularx}
\usepackage{iccc}

\usepackage{times}
\usepackage{helvet}
\usepackage{courier}

\DeclareFloatingEnvironment[%
    listname={List of Prompts},
    name=Prompt,
    placement=tbh,
    within=section
]{prompt}
\DeclareFloatingEnvironment[%
    listname={List of Stories},
    name=Story,
    placement=ht,
    within=section
]{story}

\newcommand\phm{$\phantom{-}$}
\newcommand\phs{$\phantom{*}$}

\newcommand{\citet}[1]{%
    \citeauthor{#1}~\shortcite{#1}%
}
\makeatletter
\newcommand{\citep}[2][]{%
    \@ifempty{#1}{\cite{#2}}{(#1~\citeauthor{#2}~\citeyear{#2})}%
}
\makeatother

\newcommand{\hairspace}[0]{\hspace{.08333em}}
\newcommand{\betahat}[0]{\raisebox{-1pt}{$\hat{\beta}$}}

\title{Is Temperature the Creativity Parameter of Large Language Models?}
\author{%
    Max Peeperkorn,\!\textsuperscript{1}
    Tom Kouwenhoven,\!\textsuperscript{2}
    Dan Brown,\!\textsuperscript{3} and
    Anna Jordanous\textsuperscript{1}\vspace{.3em}\\
    \textsuperscript{1}School of Computing, University of Kent, United Kingdom\\
    \textsuperscript{2}Leiden Institute of Advanced Computer Science, Universiteit Leiden, Netherlands\\
    \textsuperscript{3}Cheriton School of Computer Science, University of Waterloo, Canada\vspace{.3em}\\
    m.peeperkorn@kent.ac.uk,
    t.kouwenhoven@liacs.leidenuniv.nl, 
    dan.brown@uwaterloo.ca, 
    a.k.jordanous@kent.ac.uk}

\begin{document} 
\maketitle
\begin{abstract}
\begin{quote}
Large language models (LLMs) are applied to all sorts of creative tasks, and their outputs vary from beautiful, to peculiar, to pastiche, into plain plagiarism.
The temperature parameter of an LLM regulates the amount of randomness, leading to more diverse outputs; therefore, it is often claimed to be the creativity parameter.
Here, we investigate this claim using a narrative generation task with a predetermined fixed context, model and prompt. 
Specifically, we present an empirical analysis of the LLM output for different temperature values using four necessary conditions for creativity in narrative generation: novelty, typicality, cohesion, and coherence.
We find that temperature is weakly correlated with novelty, and unsurprisingly, moderately correlated with incoherence, but there is no relationship with either cohesion or typicality.
However, the influence of temperature on creativity is far more nuanced and weak than suggested by the ``creativity parameter'' claim; overall results suggest that the LLM generates slightly more novel outputs as temperatures get higher.
Finally, we discuss ideas to allow more controlled LLM creativity, rather than relying on chance via changing the temperature parameter.
\end{quote}
\end{abstract}

\section{Introduction}
Large language models (LLMs), in particular instruction-tuned variants, like ChatGPT and Claude, have become ubiquitous tools in society. 
Unsurprisingly, they've also found use for performing and assisting in creative tasks, such as writing stories \cite{Calderwood2020}, poems \cite{Sawicki2023}, jokes \cite{Toplyn2022}, dialogues in video games \cite{Volum2022}, and more. 
For creative tasks, \textit{temperature} is often described as the parameter that enables creative behaviour in a language model \cite{Manjavacas2017,RoemmeleGordon2018,ChenDing2023}.
Temperature controls the uncertainty or randomness in the generation process, leading to more diverse outcomes by balancing probabilities for candidate words.
However, randomness alone does not capture a complex phenomenon such as creativity \cite{Simonton2023}, it would imply that noise is the most interesting thing one can create \cite{ZenilDelahayeGaucherel2012}.
Creativity is a quality that cannot be attributed by simply producing more diverse output but involves multiple dimensions, such as social interactions and communication, or independence and freedom \cite{Jordanous2012}, that are not easily quantifiable.
Yet, one probe into LLM creativity has found a limited positive effect on creativity at higher temperatures on a divergent association task \cite{ChenDing2023}; however, they do not account for the influence of other factors, such as different prompts.
Temperature may have an effect on creativity, simply because without variation, nothing new can be created, but it is unclear if LLMs have sufficient ``knowledge'' so that randomness is all you need.

The main goal of this paper is to investigate how temperature affects the creativity of stories.
LLMs are challenging to reliably probe and most analyses of their capabilities and behaviours involve large benchmarks, which are challenging to interpret due to their scale and aggregation of outcomes.
For example, many LLMs appear to have difficulty with understanding multiple choice questions, even though this task is part of many benchmarks \cite{KhatunBrown2024}. 
Reproducibility and reliability is further complicated by small differences in prompts that can lead to vastly different outcomes \cite{Weber2023}.
Moreover, we often do not know the training data, have insight of the model architecture and additional plugins or processing for commercial LLMs.
To improve reliability and reproducibility, we predetermine a \textit{fixed} context: model and prompt.
Key to our evaluation approach is the \textit{exemplar object}, the greedy sample for that context, which is the baseline for our experiments.

In this paper, we present an empirical analysis and a creativity evaluation with human participants of stories generated by \textsc{Llama 2-Chat} \cite{Touvron2023} on four necessary conditions for creativity: novelty, typicality, cohesion, and coherence.
The primary findings of this work are:
\begin{itemize}
\item In general, temperature does not allow the LLM to leverage different regions of the embedding space, but it does enable some novelty when generating limited samples (as is the case for any real-world application).
\item We observe a weak positive correlation between temperature and novelty, and unsurprisingly, a negative correlation between temperature and coherence. 
Suggesting a trade-off between novelty and coherence.
\end{itemize}

Overall, the influence of temperature on creativity is far more nuanced and weak than the ``creativity parameter'' claim suggests.
Finally, we discuss the limitations of our approach, and conclude with a discussion on creative outputs of LLMs to rely less on chance.

\section{Background}\label{sec:background}
The generative aspect of LLMs and other large generative models naturally sparked an interest for application towards creative tasks. 
The jump in output fidelity and ease-of-use are proving to radically change the landscape of creative tasks that a machine can perform.
LLMs certainly seem capable of producing novelty, typicality, and grammatically fluent output \cite{PeeperkornBrownJordanous2023}, but it is difficult to determine how exactly to approach such a creativity evaluation \cite{LambBrownClarke2018}. 
In this section, we describe the difficulties with probing LLMs and build a foundation drawing from cognitive science for our research approach.
To investigate temperature as the creativity parameter, we first need to establish what we mean by creativity and what conditions we require outputs of the LLM to satisfy.

\subsection{Conditions for Creativity in Narrative Generation}
In artistic domains, we consider something creative if it is novel or original and useful or effective \cite{Boden1992,RuncoJaeger2012}.
Usefulness is closely related to quality and value, as things of low-quality tend not to be very useful or have much value.
Moreover, because creativity is such a complex phenomenon, further properties have been proposed, such as surprise \cite{GraceMaher2014,Simonton2012} and typicality \cite{Ritchie2007}. 
Both of these properties are often viewed as variants of novelty, and the differences are very subtle. 
If something surprises you, there must be something that you have not seen before, something unexpected. 
Indeed, \citet{Martindale1990} draws a similar observation, noting that novelty is the disruption of expectedness.
Typicality is more interesting because the difference is that before we ask any questions regarding novelty and quality, we must consider the prior question; to what extent is something typical? 
Is the object an example of its class or genre? 
Typicality and novelty are, like surprise and novelty, somewhat at odds, as high typicality often implies low novelty; however, Ritchie suggests that typicality is a prerequisite for evaluating creativity, meaning that for something to be judged as novel it has to some extent be typical first.
This view lends itself well for this work because we only consider the creativity of LLMs based on the artefacts it produces and not in a larger scope.
While we do not use Ritchie's empirical criteria directly, his perspective fits our purpose as we mainly evaluate the artefact of the system and not the process behind it, over a framework like the Creative Tripod \cite{Colton2008}, which specifically focuses on the behaviour of creative systems.
\citet{Ritchie2007} formulates the essential properties as follows:
\textit{novelty} is the extent to which the produced item is dissimilar to existing samples within its class, \textit{typicality} is the extent to which the produced item is an example of the class in question, and \textit{quality} is the extent to which the produced item is a high-quality example of its class.

For narrative writing tasks specifically, it is relatively poorly understood what is valued for creativity, for example, the importance of genre is unclear \cite{DSouza2021}.
In a large creative writing literature survey, \citet{DSouza2021} outlines various criteria drawn from a large body of work. For example, a narrative needs to have an elaborate plot, an original voice, interesting characters, a good flow, linguistic originality, among many more. 
These criteria largely connect with the essential properties above, and we use these to outline four necessary conditions to establish if temperature affects creativity for the narrative generation task in this study.
Hence, we require novelty, typicality, and quality in terms of cohesion and coherence \cite{Graesser2004}.

Novelty and typicality are relatively straightforward to reformulate in this domain.
We would say that a story is \textit{novel} if it is dissimilar to other stories, if it has some characteristics that set it apart from other stories, for example, it may be written in a novel setting or narrative style.
A story is considered \textit{typical} if it is of the same form as other stories in the same category. 
For example, based on stories we have read before, we can observe certain patterns, narrative style, and other characteristics.
When we are presented with another story, we say it is typical if it follows similar rules or patterns and has those same characteristics, i.e. it has the same form.

Quality, in general, can mean a wide variety of things, and can be defined depending on what you value.
For example, large language models can easily and fluently write high-quality texts in terms of grammar and spelling, but this falls short for our purposes since the generated stories can be entirely meaningless or nonsensical. 
A key aspect of a good story is a good narrative, that it is cohesive and coherent. 
The difference between cohesion and coherence is subtle, as pointed out by \citet{Graesser2004}.
Cohesion is about the objective features in a text that hold everything together, in other words, sentences are well-connected, are characters, events, and actions used appropriately. 
Coherence, however, is about the reader's interpretation and the meaningfulness of the text. 
The reader should be able to easily follow and understand what it means. 
For our purposes, we consider the quality of a story as a combination of cohesion and coherence; 
A story is considered \textit{cohesive} if words and sentences are connected and grammatically consistent, and that characters, events and actions are used appropriately. A story is \textit{coherent} if it is easy to follow the story and understand what it is about.

\subsection{Probing Large Language Models}
There are roughly three techniques in which we can ``control'' the LLM; through training data (either from scratch or fine-tuning), via in-context learning \cite{Brown2020}, and with different hyperparameter configurations.
We focus here on using LLMs that are already instruction-tuned, meaning they are fine-tuned for conversation following a specific prompting format. 
An instruction-tuned LLM can respond to instruction and appear to ``understand'' what is asked of them \cite{Brown2020}.
This suggests that by changing the context in the prompt, we can tap into different ``slices of knowledge'' of the learned probability distribution.
In a co-creative or interactive setting, one might ask to change details and refine previously generated poems or stories, allowing for more creative outcomes.
In essence, by tweaking the prompt, we narrow down the sampling distribution closer to the intended outcome.
Adjusting the prompt to get the ``right'' slice of the probability distribution, however, is very sensitive to the choice of words and even punctuation \cite{Weber2023}. 
It becomes increasingly difficult to separate the effect of the prompt as you allow more randomness via the temperature hyperparameter in the generation process.

If temperature is the creativity parameter, then you would expect something similar to what happens when you change the prompt; that at higher temperatures, the LLM will venture into different slices of its knowledge.
However, it is difficult with this kind of investigation to reliably reproduce outputs and make comparisons across different models, prompts and hyperparameter settings, to draw any conclusions.
Therefore, it is essential for our evaluation to fix the context; the prompt and model, and all other parameters, except for temperature.

\subsection{The Temperature Parameter}
Temperature is a hyperparameter $t$ that we find in stochastic models to regulate the randomness in a sampling process \cite{Ackley1985}. 
The softmax function (\autoref{eq:softmax}) applies a non-linear transformation to the output logits of the network, turning it into a probability distribution (i.e. they sum to 1). The temperature parameter regulates its shape, redistributing the output probability mass, flattening the distribution proportional to the chosen temperature.
This means that for $t > \textrm{1}$, high probabilities are decreased, while low probabilities are increased, and vice versa for $t < \textrm{1}$.
Higher temperatures increase entropy and perplexity, leading to more randomness and uncertainty in the generative process. 
Typically, values for $t$ are in the range of $[\textrm{0, 2}]$ and $t = \textrm{0}$, in practice, means greedy sampling, i.e. always taking the token with the highest probability.

\begin{equation}\label{eq:softmax}
    \mathrm{softmax}(\mathbf{z})_i = \frac{%
        \exp(\frac{z_i}{t})
    }{%
        \sum^n_{j} \exp(\frac{z_j}{t})
    }\quad\mathrm{where}\ \mathbf{z} \in \mathbb{R}^n
\end{equation}

As more randomness and uncertainty is introduced in the generation process, the more difficult it becomes to probe LLMs and compare outputs.
To investigate the effects of temperature, we need a consistent baseline, for which we propose to use the greedy sample.
In the next section, we reason why the greedy sample is a suitable point of departure.

\subsection{Prototypes and Exemplars}
The method for evaluation we propose in this paper is inspired by two theories of cognitive categorisation, namely prototype theory \cite{Rosch1973} and exemplar theory \cite{MedinSchaffer1978}.
These two theories are often contrasted, but have much overlap. 
The key difference is that in exemplar theory, categorisation of new stimuli is not compared against a single prototype object, but against multiple exemplars of the category.
A prototype is considered something that is derived from presented stimuli, where an exemplar is one of the previous stimuli.
Prototype theory observes that many ideas and objects can be put in the same cognitive category based on their name, but as one looks closer may not have that many things in common.  
This relates strongly to the idea of \textit{family resemblances} described by Wittgenstein~\shortcite{Wittgenstein1953}, where he argues that even though items may have similar names, there is not necessarily one thing common to all, but a complex network of similarities and relationships between members of the category.
In a category, basic objects (prototypes) are the ones that carry the most information, are most representative of their category, and their information can be used as a compressed representation of other class members.
\citet{Rosch1976} show that basic objects are first categorisations made for that category of objects in an environment. 
This is reminiscent of exemplar theory, and people often use a mixture of prototypes and exemplars as they categorise objects.
When people are primed with just exemplars, they can still accurately describe the prototype \cite{MedinAltomMurphy1984}.

These theories have further connection with conceptual spaces, geometric mental structures, where each point represents an object, the dimensions are properties, and concepts (categories) are convex regions \cite{Gardenfors2014}. 
The central member, the prototype, or most typical object, has the most in common with most objects in the category, i.e. it has the short average distance to all other objects in the category.
This does not mean that it has overlap with all other objects, but has some resemblances with each member of the category.
Conceptual spaces also play an important role in Boden's framework of creativity \cite{Boden1992}. 
She uses a metaphor, a map of the mind, to describe exploratory creativity and that novelty and value can be found venturing into unknown regions.
To be more precise, the typicality of an object relates to how well it fits with the exemplars or the prototype, while novelty could be viewed as how much the category changes or expands as a new stimulus is presented.
This implies similarly that if temperature is indeed the creativity parameter, then it should allow access to different slices of the probability distribution or other regions in the embedding space.
Prototype and exemplar theory provide a useful frame that motivates our approach to evaluation, and a method to evaluate further stimuli relative to a baseline (the central member or prototype). Next, we delve into what the exemplar is and how to generate it with LLMs in mind.

\section{The Exemplar Story}\label{sec:exemplar-story}
Rigorously evaluating LLMs is hard.
If we consider multiple models, it is difficult to compare their results because the learned representations and the training data could be very different.
In the case of a single model, but many prompts, it is difficult to establish a solid baseline, and therefore also impossible to compare results, since lexical changes can have a cascading effect on the output.
Hence, we need to work within a \textit{fixed} context using a specific model and prompt to establish a baseline that enables the comparison of results against independent variables, such as temperature.

\subsection{Identifying the Exemplar}
In the \nameref{sec:background} section, we point out that prototype/exemplar theory is a useful frame for a comparative evaluation, but what object can we consider as the exemplar or prototype? 
As mentioned, we require a baseline and a fixed context to mitigate some difficulties of probing and instructing LLMs, and there are two perspectives to determine what we consider the prototype or exemplar object.

First, seeing that the peak of distributions is the most probable or average outcome, using the prototype theory perspective, it is straightforward to view the greedy sample as the prototype.
In that case, a category is every possible output produced by one prompt and one model, with the greedy sample as its prototype.
However, there are many prompts that can generate similar outcomes.
The second perspective departs from this more abstract view.
In that frame, the category is a group of prompts and a model, and the greedy sample for each prompt is an exemplar.
For the purposes of this work, we call the greedy output the \textit{exemplar object}, but notably, prototype would also be a viable description, since we consider only one prompt for the human evaluation.

\begin{prompt}[!ht]
\small
\begin{lstlisting}
[INST]Write a story.[/INST]Here it is:\n\n
\end{lstlisting}
\caption{The exact \textsc{Llama 2-Chat} prompt, showing the chat formatting and the suffix to make sure the model starts writing the story immediately.}\label{fig:exact-llama-2-prompt}
\end{prompt}

\subsection{Generating the Exemplar Story}
We generate the exemplar and all other stories, using the instruction-tuned \textsc{Llama 2-Chat} 70B model.
This autoregressive LLM is competitive, but more importantly, it is open source,\!\footnote{The \textsc{Llama 2} licence restricts commercial use, which strictly speaking is not open source as defined by the Open Source Initiative.} so that we have full access to the model architecture and weights.
This is essential since we aim to eliminate or minimise any effects besides the temperature parameter.
Closed-source models, like GPT-3.5 and GPT-4, are for these reasons not viable as they are not transparent.
\textsc{Llama 2} is trained on an (unspecified) mix of publicly available data, and the chat variants are fine-tuned using reward modelling on annotated human preference data \cite{Touvron2023}.

For the instructions, we aimed for a minimal and neutral prompt (\autoref{fig:exact-llama-2-prompt}) so that it introduces as little additional information as possible.
We applied 6-bit quantisation,\!\footnote{Quantised using the llama.cpp library with setting Q6\_K. See \url{https://github.com/ggerganov/llama.cpp/pull/1684}} which reduces the model size with extremely minimal loss of quality \cite{Zhang2023}, to make it fit on available hardware.
\textsc{Llama 2-Chat} 70B adds, unsurprisingly, much conversation to the output, instead of just performing the task, even when instructed to avoid this in the system prompt. 
This is unwanted behaviour, as it has a large effect on subsequent stories.
To mitigate this, we added a minimal start of the LLMs response (see \autoref{fig:exact-llama-2-prompt}) to make sure that this is consistent across all generated stories. 
Another benefit is that it reduced any post-processing for the human evaluation study, besides, occasionally, pruning one sentence at the end of the output that was not part of the story.
We set top-$k$~=~50 and disabled all other decoding strategies. 
Setting top-$k$ and thereby limiting the token candidates at each time step, influences the output but is necessary; otherwise, text quality quickly deteriorates \cite{Holtzman2020}, especially at higher temperatures.
Next, we present a case study consisting of both a computational analysis and human evaluation to approach the research question. 

\begin{figure*}[t]
    \centering
    \begin{subfigure}[t]{0.32\textwidth}
        \centering
        \includegraphics[width=\textwidth]{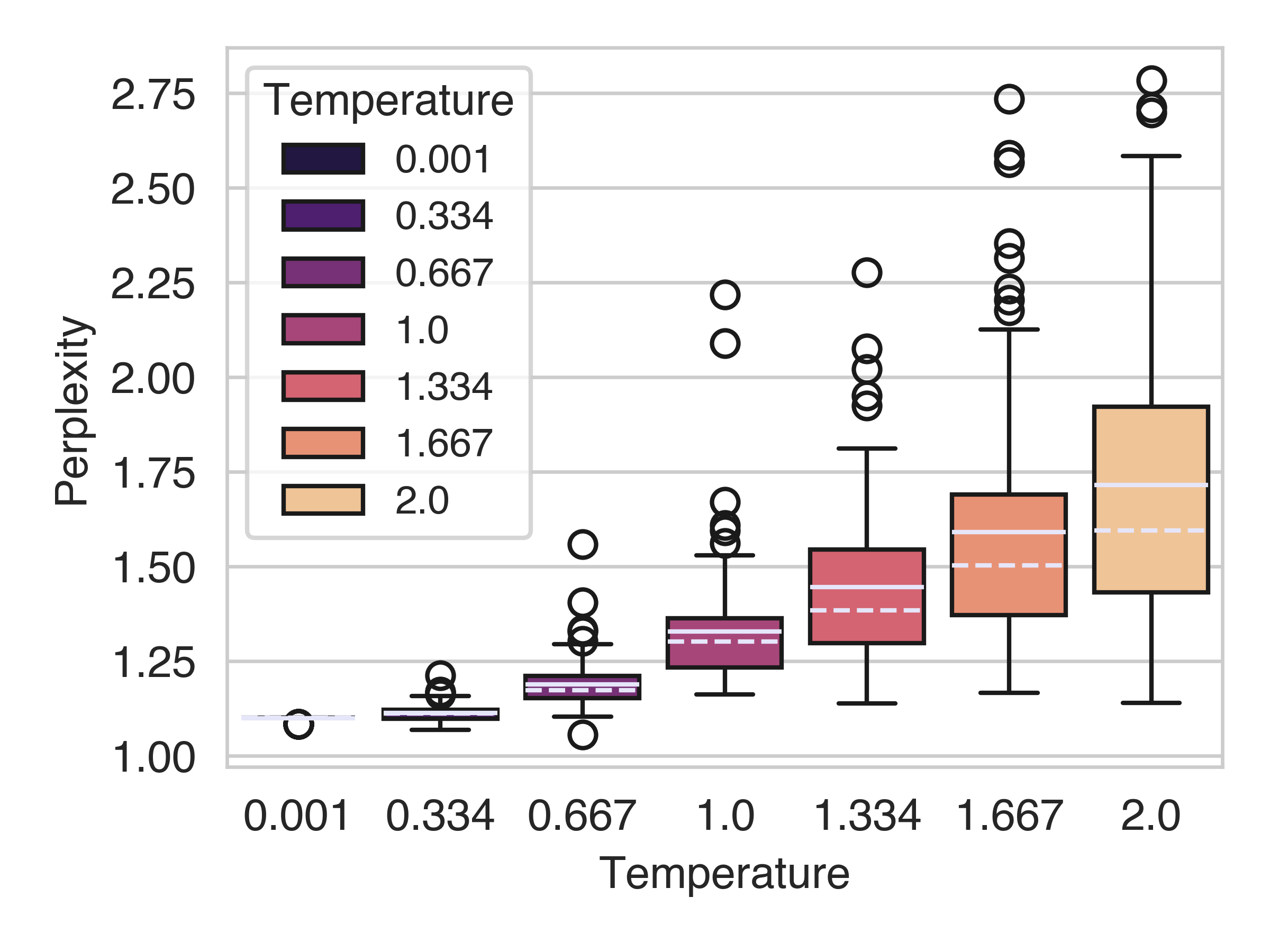}
        \caption{Perplexity of stories given the LLM per temperature value.}
        \label{fig:comp-eval-ppl}
    \end{subfigure} 
    \hfill
    \begin{subfigure}[t]{0.32\textwidth}
        \centering
        \includegraphics[width=\textwidth]{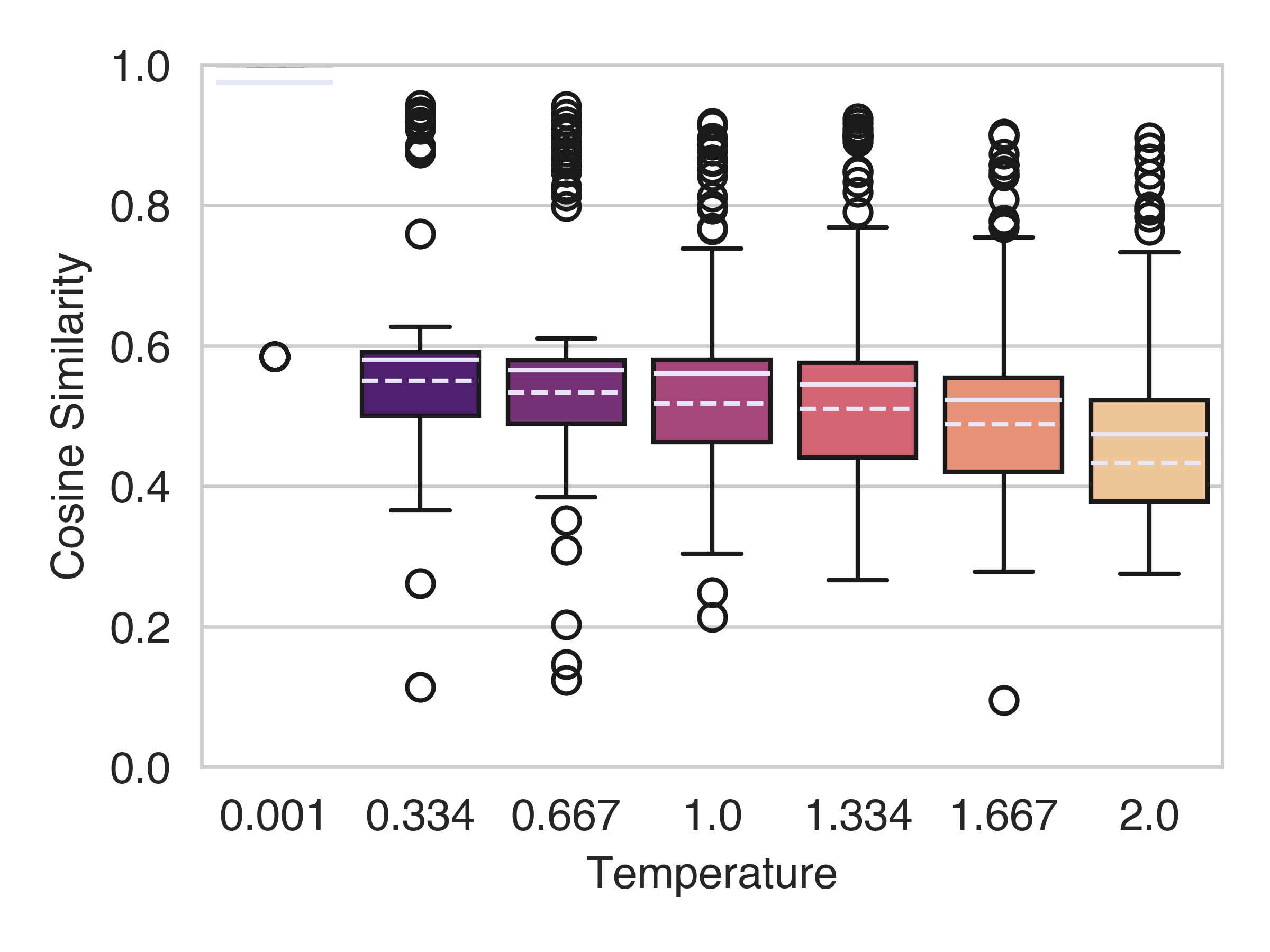}
        \caption{Cosine similarity of stories with the exemplar per temperature value.}
        \label{fig:comp-eval-cossim}
    \end{subfigure}
    \hfill
    \begin{subfigure}[t]{0.32\textwidth}
        \centering
        \includegraphics[width=\textwidth]{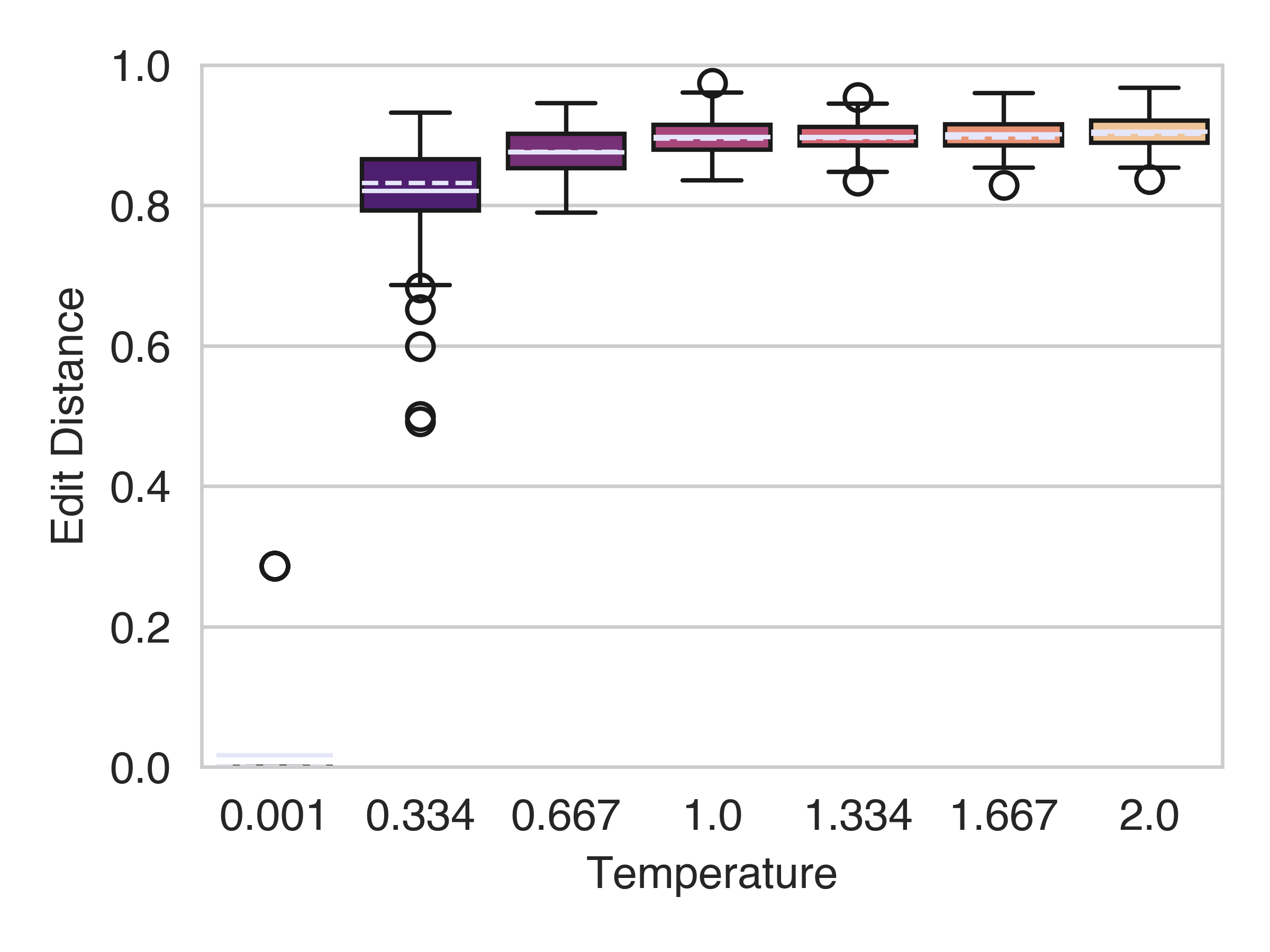}
        \caption{Normalised edit distance of stories with the exemplar per temperature value.}
        \label{fig:comp-eval-edit}
    \end{subfigure}
    \caption{In (a), we see that the effect temperature has on the perplexity, the quality according to the LLM, of the output. Both, (b) and (c) suggest that higher temperatures do not imply more diversity on the semantic or lexical level, or that it further extends the range of possible outputs in the current context.}
    \label{fig:comp-eval}
\end{figure*} 

\section{Case Study -- ``Write a story.''}
Our approach to investigating how temperature affects creativity in LLM-generated stories is a two-fold empirical analysis.
First, we perform a computational analysis to examine the behaviour of LLMs for different temperature values. In particular, how the stories are distributed in the embedding space and measure distance against the exemplar object.
Secondly, we present an experimental study in which humans evaluate creativity to investigate the interaction between temperature and the four necessary conditions of creativity for a narrative generation task.

\subsection{Computational Analysis}\label{ssec:comp-analysis}
The goal of this analysis is to show if temperature enables the LLM to access slices in its probability distribution or regions in its embedding much further than those sampled at lower temperatures.
We do not aim to evaluate stories for their creativity using the four necessary conditions, but establish any evidence from a computational perspective for the possibility of any exploratory creative behaviour.
With that in mind, \citet{Boden1992} argues, using conceptual spaces, that exploration is essential for creative behaviour.
Moreover, an increased diversity of outputs might suggest an increased likelihood of producing something novel.
For the computational analysis, we use the same setup as described above, using \autoref{fig:exact-llama-2-prompt} and its exemplar object (see \nameref{supplementary-materials}).
We generate 100 stories over 7 different temperature values $t \in [\hairspace\textrm{.001, .334, .667, 1.0, 1.334, 1.667, 2.0}\hairspace]$, to investigate the effect of temperature on the diversity of stories.

\subsubsection{Evaluation Measures}
In general, the quality of LLM-generated text is usually assessed using perplexity. 
This could be done with the same model, however, for purposes of this study, this does not inform much, as at higher temperatures the perplexity on average goes up (\autoref{fig:comp-eval-ppl}), meaning that the quality goes down.
Measuring quality using an independent model is more appropriate, but it raises the question of what is a suitable candidate for a narrative generation task.
Moreover, previous work suggests that perplexity is a bad indicator of text quality compared to human evaluations \cite{Zhang2021}. 
For these reasons, we focus more on diversity and similarity.
An overall higher diversity proportional to temperature should indicate that different slices of the probability distribution are used to generate the outputs.
Comparisons for diversity and similarity can be done on the lexical level, which uses the texts directly, or on a semantic level using stories embedded in the LLM's embedding space.
We focus on two commonly used measures that can be used relative to the exemplar; on the semantic level, we use cosine similarity with the embeddings of the stories and the exemplar, and for the lexical level, we use normalised edit distance between stories and the exemplar.
If temperature is proportional to creativity, then you would expect the outputs to extend into very different regions of the embedding space.
We examine the distribution using principal component analysis (\textsc{pca}), preserving global structure, to project embeddings of the generated stories in two-dimensional space.

\subsubsection{Results}
In general, the distribution plots (\autoref{fig:comp-eval}) suggest that the LLM, for this specific prompt and model, is unable to produce more diversity at higher temperature.
Higher temperatures increase the odds of generating that diversity, but the measures indicate that it is not essential.
In \autoref{fig:comp-eval-edit}, the normalised edit distance measure shows that even at lower temperatures ($\textrm{.334} < t < \textrm{1.0}$), there is an immediate effect on diversity relative to the exemplar story. 
Although, it does not tell anything about the meaning of the stories, it suggests that higher temperatures ($t > \textrm{1.0}$) do not necessarily lead to more diversity.
Cosine similarity shows a weak negative trend, but also shows that there is great overlap in the range of outputs for each temperature value (\autoref{fig:comp-eval-cossim}). 
In other words, outputs at higher temperatures on the semantic level, do not extend much further away from the exemplar story than outputs at lower temperatures.

\begin{figure*}[!ht]
    \centering
    \includegraphics[width=\linewidth]{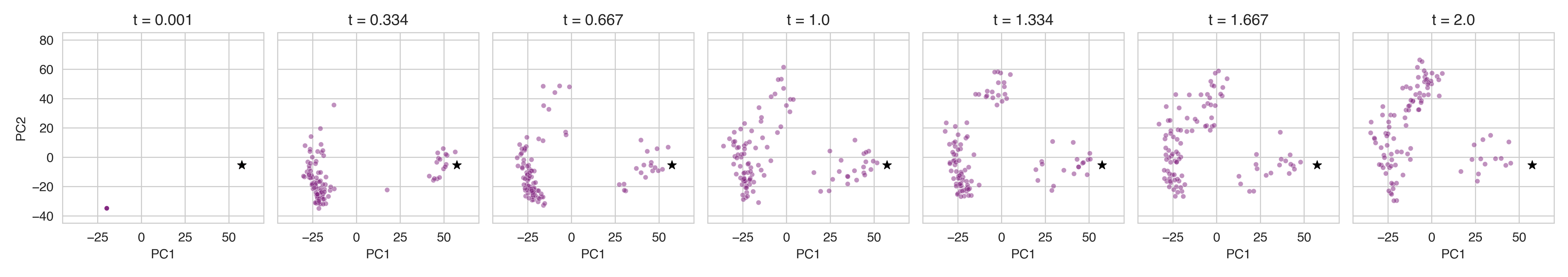}
    \caption{Here, we show the \textsc{pca} projections of 100 stories per temperature generated using \textsc{Llama 2-Chat} 70B and \autoref{fig:exact-llama-2-prompt}. 
    While we observe that an increasing temperature seems to explore a larger region of the embedding space with a small number of samples, it does not imply that we access a broader slice of the model's probability distribution. We merely increase the chance of generating more diversity.
    $\bigstar$ denotes the exemplar story.}
    \label{fig:pca-embeddings-write-per-temperature}
\end{figure*}

In theory, given enough time and space, we should be able to generate outputs from every region of the embedding that the context of prompt and model can be generated, even at lower temperatures.
However, when drawing limited samples from the LLM, as is standard in any real-world scenario, we do observe some exploration at higher temperatures as is shown in \autoref{fig:pca-embeddings-write-per-temperature}.
At the very least, it appears that temperature increases the chance of finding novelty more quickly.

Another interesting observation is that it seems, relative to the exemplar story, that the generation process moves in a particular direction (\autoref{fig:pca-embeddings-write-per-temperature}). 
This suggests some merit to the link we draw with prototype and exemplar theory and our approach to evaluation, as it shows how LLM outputs changes relative to the exemplar.
Informally, we observed similar behaviour for seemingly synonymous prompts, this might warrant further investigation in future work.

\subsection{Experimental Analysis with Human Evaluation}
The computational analysis is limited because it does not inform about the meaning of the stories.
It might be that even though our measures show clear limitations in terms of diversity and overlap in output distributions, stories generated at higher temperatures are indeed creative in ways not captured by the computational analysis.
To further investigate this, we designed a creativity evaluation experiment with non-expert human participants.
In the experiment, participants evaluate stories according to the four necessary conditions outlined in the \nameref{sec:background} section. 
Specifically, we ask them to compare each story for novelty and typicality against the exemplar and judge the story for its individual cohesion and coherence.
To generate the stories, we again use \textsc{Llama 2-Chat} 70B with the same prompt (\autoref{fig:exact-llama-2-prompt}), with the same setup described in the Exemplar Story section.
For each of the 7 temperature values $t \in [\hairspace\textrm{.001, .334, .667, 1.0, 1.334, 1.667, 2.0}\hairspace]$, we randomly generate 5 unique stories that are at least 300 tokens long.
At the lowest temperature setting $t$~=~.001, the model could generate only one other unique story beside the exemplar story, as such we have a total of 31 stories for the creativity evaluation.

\subsubsection{Experimental Setup}
For this study, we recruited 36 participants online and from the Computational Creativity module at the University of Kent. 
Each participant was asked to evaluate 5 LLM-generated stories using a questionnaire (see \nameref{supplementary-materials}). 
They were told upfront that the stories were generated by a LLM, but they were unaware of the specific model or that the stories were sampled from different temperature values.
The participants were given the definitions of the four necessary conditions for creativity, followed by the exemplar story, and a two-story practice evaluation.
The practice stories are the same for all participants and given in random order.
After practice, they were again presented with the exemplar story before starting the actual creativity evaluation.
The exemplar story and definitions were accessible throughout the survey.
The study was approved by the Central Research Ethics Advisory Group of the University of Kent.

Each participant was randomly assigned 5 of the 31 stories.
The evaluation consists of 4 five-point scale questions; two about novelty and typicality \textit{compared against the exemplar story}, and two about the stories' text cohesion and coherence. 
The questions about novelty, typicality, and cohesion are on a five-point scale from 1) not at all \underline{\ \ \ \ \ }, to 5) completely \underline{\ \ \ \ \ }.
Notably, coherence is on a five-point scale from 1) very easy, to 5) very difficult to understand and follow.
As a result, lower scores indicate high coherence, and vice versa.
The survey distribution procedure was set up such that each story received at least five evaluations by different participants.

Concluding the survey, we posed 5 post-evaluation questions about their experience with reading, writing, and reviewing stories, their stance towards Generative AI and its use for creative tasks, and how they approached the evaluations.

\subsubsection{Practical Considerations}
There are several practical considerations regarding the experimental design that limited the study in various ways. 
The key limitation of this study is its size; we recruited 36 participants, who can only evaluate a limited number of stories.
We generated only 5 unique stories per temperature, relying on the sampling distribution of the LLM to provide an appropriate sample. 
Ideally, you would generate many stories and then take a representative sample, but we do not know what number is sufficient for our purpose.
Similarly, we only considered one prompt and its exemplar, ultimately, we decided on, in our opinion, the most minimal and neutral instruction to perform the task (\autoref{fig:exact-llama-2-prompt}).

In this kind of setup, there is a risk of practice and order effects. 
To mitigate these issues, we asked the participant to first do a practice evaluation, making sure that the participant is fully aware of the task and what is expected of them.
As mentioned, we randomly assigned stories to the participants, and did not observe any ordering effects in the analysis.
We considered using pairwise comparison ranking, and presenting stories in pairs and evaluating those relative to each other, to further mitigate these effects.
While this method is effective towards that purpose, it does not scale very well, even with just 31 stories, it requires a large number of participants for robust and meaningful outcomes.

Another limitation of this study is that cohesion and coherence are, in fact, important factors for novelty and typicality.
We did not observe any issues with that in the analysis, and as these LLMs produce fluent and grammatically correct language, this does not appear to be an issue.
The stories generated by \textsc{Llama 2-Chat} 70B are simple stories featuring only basic elements, and are not literary masterpieces. 
We are interested in a relative evaluation to see how temperature changes the creativity as it increases only for a simple task; thus we opted to use non-experts who are not preconditioned to higher expectations.
For that same reason, we designed the survey around specific criteria and did not ask participants directly about the creativity of the stories.

\subsubsection{Analysis}
We use linear mixed-effects models \cite{SeaboldPerktold2010} to control for the random effect of assigning stories to different participants, and Cronbach's $\alpha$ to test inter-rater reliability.
The approach is as follows:
First, we determine if there is a significant effect using Satterthwaite's method \cite{Satterthwaite1941}. 
Secondly, the model's estimate \smash{\betahat} (slope) determines the direction of the effect, i.e. positive or negative, and the rate of change.
Finally, we test how much variance is explained by the model to determine the magnitude using conditional \smash{$R^2$} \cite{NakagawaSchielzeth2013}, denoted by \smash{$R^2_c$}, which considers both fixed and random effects.
The more variance is captured by the linear mixed-effects model, the stronger the correlation.
Additionally, we report the marginal \smash{$R^2_m$}, which is the variance explained by the fixed effects.

\begin{table}[!t]
\centering
\small
\renewcommand{\arraystretch}{1.2}
\caption{Evaluation Ratings and Reliability}\label{tab:desc-stats}
\begin{tabularx}{\linewidth}{X|cc|c}
    \toprule
    & Mean & Std. Dev. & Cronbach's $\alpha$ \\
    \midrule
    Novelty    & 3.12 & 1.17 & .547 \\
    Typicality & 3.20 & .987 & .664 \\
    Cohesion   & 3.69 & .959 & .596 \\
    Coherence\textsuperscript{$\dagger$} & 2.13  & 1.17 & .749 \\
    \midrule
    Mean      & \multicolumn{2}{c|}{---} & .639 \\
    \bottomrule
    \multicolumn{4}{r}{{\scriptsize $\dagger$ lower indicates higher coherence}}
\end{tabularx}
\end{table}

\begin{table*}[!ht]
\centering
\small
\renewcommand{\arraystretch}{1.2}
\addtolength{\tabcolsep}{-1pt}
\caption{Statistical Analysis of Creativity Evaluation and Computational Metrics}\label{tab:eval-and-metrics+r2m}
\begin{tabularx}{\linewidth}{X|@{}ccc|@{}ccc|@{}ccc|@{}ccc}
    \toprule
     & \multicolumn{3}{c|}{Temperature} & \multicolumn{3}{c|}{Perplexity} & \multicolumn{3}{c|}{Cosine Similarity} & \multicolumn{3}{c}{Normalised Edit Distance} \\
     & $\hat{\beta}\pm\mathrm{SE}$ & $R^2_c$ & $R^2_m$ & $\hat{\beta}\pm\mathrm{SE}$ & $R^2_c$ & $R^2_m$ & $\hat{\beta}\pm\mathrm{SE}$ & $R^2_c$ & $R^2_m$ & $\hat{\beta}\pm\mathrm{SE}$ & $R^2_c$ & $R^2_m$ \\
    \midrule
    Novelty    & \phm.308$\pm$.138* & .385 & .152 & \phm.730$\pm$.273** & .370 & .158 & -1.08$\pm$.913\phs   & .338 & .129 & \phm2.01$\pm$.731** & .378 & .161 \\
    Typicality & -.095$\pm$.118     & .339 & .125 & -.205$\pm$.237\phs  & .345 & .126 & -.987$\pm$.771\phs   & .336 & .128 & -.662$\pm$.632\phs & .336 & .126 \\
    Cohesion   & -.181$\pm$.112     & .406 & .161 & -.183$\pm$.226\phs  & .392 & .151 & \phm2.09$\pm$.716** & .431 & .187 & -.663$\pm$.608\phs & .384 & .152 \\
    Coherence\textsuperscript{$\dagger$}  & \phm.240$\pm$.122* & .646 & .283 & \phm.194$\pm$.251\phs\phs   & .627 & .274 & \phm1.03$\pm$.788\phs\phs   & .628 & .278 & \phm.750$\pm$.664\phs\phs & .629 & .276 \\
    \bottomrule
    \multicolumn{13}{r}{\raisebox{-2.5pt}{*} {\scriptsize $p < \textrm{.05}$}, \raisebox{-2.5pt}{**} {\scriptsize $p < \textrm{.01}$}, {\scriptsize $\dagger$ positive correlation denotes a negative effect, i.e. higher temperatures are less coherent}}
\end{tabularx}
\end{table*}

\subsubsection{Quantitative Results}
The primary results of the analysis focus on the effect of temperature on the creativity of stories. 
Regarding inter-rater reliability, on average, we find $\alpha$~=~.639 (\autoref{tab:desc-stats}), which is just short of the acceptable level of .7, but still indicates some degree of agreement, more so because we are dealing with non-expert judges who, in general, disagree more often.
Regarding temperature, we observe two effects (\autoref{tab:eval-and-metrics+r2m}), a weak positive correlation between novelty and temperature (\smash\betahat~=~.308, SE~=~.138, \smash{$R^2_c$}~=~.385), and a moderate correlation between coherence and temperature (\smash\betahat~=~.240, SE~=~.122, \smash{$R^2_c$}~=~.646).
This implies a trade-off between novelty and coherence.
While higher temperatures seem to positively affect the chance of finding novel stories, they become less coherent.
Although, coherence is rated lower (M~=~2.13), meaning that the stories, overall, were easy to understand.
The negative correlation of coherence with temperature is as expected \cite{Holtzman2020}. 
Both effects, have relatively shallow slopes, indicating that the rate at which novelty and coherence change as temperatures get higher is low, indicating only a small effect.

The secondary results relate the creativity evaluation to the computational metrics in the \nameref{ssec:comp-analysis} section.
Note that perplexity is a metric relative to the model, and cosine similarity and normalised edit distance are both relative to the exemplar story.
Here, we observe three effects for perplexity, cosine similarity and normalised edit distance in \autoref{tab:eval-and-metrics+r2m}. 
First, we see that perplexity has a weak to moderate positive correlation with novelty (\smash{\betahat}~=~.730, SE~=~.273, \smash{$R^2_c$}~=~.370). 
This is interesting because it suggests that perplexity, although a measure for model quality, could also indicate novelty at least to some extent. 
Yet, it is also unsurprising because indirectly, higher perplexity is on average a result of higher temperature (see \autoref{fig:comp-eval-ppl}). 
There is likely a similar cause for the weak  to moderate positive effect between normalised edit distance and novelty (\smash{\betahat}~=~2.01, SE~=~.731, \smash{$R^2_c$}~=~.378).
Finally, there is a weak to moderate positive effect between cosine similarity and cohesion  (\smash{\betahat}~=~2.09, SE~=~.716, \smash{$R^2_c$}~=~.431), which is not unexpected as higher temperatures produce outputs with greater variety, subsequently, increasing the likelihood of producing non-cohesive texts.
More interestingly, this suggests that stories closer to the exemplar are more cohesive. 
In turn, this implies when the semantic distance between a story and the exemplar is large, more features related to cohesion are lost.
The slopes of the analysis with the computational metrics are steeper, especially the latter two have a high rate of change, looking at \autoref{fig:comp-eval-cossim} and \autoref{fig:comp-eval-edit}, this could be due to the large difference between the lowest and highest temperatures.

\subsubsection{Qualitative Results}
In the post-evaluation questions, the participants reported on average some familiarity with generative AI, and were moderately comfortable with AI being used for creative work.
Most participants reported very little to no experience with writing and reviewing stories. 
Those that did report to have done writing were mostly relying on educational experience.
We specifically asked the participants about their approach for evaluating stories to gain further insights in what specifically they focused on.
Participants reported different ways in which they approached the evaluation, some focused on structure, another aimed to grab the general idea combined and compared the ending. 
Occasionally, the participant chose one specific factor, such as most stories were adventure-based, and took that as the typical trait. 
Another example, is that when a story switches genre, for example, to fantasy, it was considered more novel.
In line with this, participants did report that it was quite difficult to separate details in the stories as they went through the evaluation, and that it was quite difficult to just use the four conditions as there were different characteristics in the story they had to weigh. 
One participant reported a lack of cohesion, and that sometimes it felt that sentences were skipped, misplaced, or there was something missing.
These reports suggest that for each criterion, different things are expected and valued in LLM-generated stories.

\section{Discussion and Further Work}
Overall, we do not find compelling support for the ``temperature as the creativity parameter'' claim across the conditions for creativity in this investigation. 
While we see some positive effects for evaluating novelty, the results show that important aspects required for creativity are missing for temperature to not enable LLM creativity to the full extent.
In this section, we first discuss some observations about our analysis and methodology and then outline three recommendations for future work.

The key idea to our evaluation methodology is the idea of the exemplar, inspired by Prototype and Exemplar theory, as a point of departure for evaluation.
As mentioned, we observe that the exemplar appears to be on the edge of the projection, while this suggests that the exemplar is a reasonable baseline.
To further examine the merits of the link we draw with prototype and exemplar as an evaluation methodology, further investigation is needed to see if this phenomenon emerges for different domains, prompts, and models.

In the qualitative analysis, the participants reported using different criteria to evaluate novelty and typicality and found it difficult to separate details between stories during the evaluation.
While this is nothing new, these reports complicate the robustness of our findings.
Repeating the experiment with experts, better stories, and creative writing rubrics, is important for future work. 
Moreover, an expert's depth of understanding of the domain could identify other more advanced desiderata.
However, we envision that it will remain difficult to design a robust, large-scale experiment to investigate the effect of temperature on creativity because of the vast amount of variation that LLMs can produce.

\subsection{Towards More Creativity in LLMs}
The inherent complexity of creativity means it is highly unlikely that there exists such an easy solution as a single parameter that enables creativity in LLMs.
Of course, careful fine-tuning for specific tasks or prompting might produce the desired outcomes, however, fine-tuning comes at the cost of generality, and prompting is unpredictable and inconsistent across different models.
From this work, we suggest future work that could be useful for progressing LLM creativity.

\subsubsection{Benchmarks for Creativity} 
In general, the creative abilities of LLMs are mostly evaluated on tests from psychology (e.g. Torrance Test of Creativity Thinking, Alternative Uses Test, Divergent Association Test), or inferred from phenomenological observations in benchmarks of other tasks, such as mathematical reasoning \citep{Bubeck2023}. 
To our knowledge, there are no strong LLM creativity benchmarks that go further. 
We only presented a minimal case and reliably scaling it up is challenging, and this is partly due to the complexity of evaluating creativity, but it is important to investigate what such benchmarks should be, to make more robust claims about the creativity of LLMs.

\subsubsection{Decoding Strategies} 
More advanced decoding strategies could be interesting if designed for specific purposes.
Decoding strategies are vital for the LLM to produce fluent natural language, and might be similarly helpful for producing creative writing or communication.
Besides the two well-known parameters, top-$k$ and nucleus sampling \cite{Holtzman2020}, there are more complex decoding strategies, namely mirostat \cite{Basu2021} and locally typical sampling \cite{Meister2023}. 
A decoding strategy designed with an information-theoretic notion of creativity \cite{MondolBrown2021} in mind might be fruitful to enable more creative behaviours regardless of model and prompt.

\subsubsection{Implicit Information} 
LLMs appear to capture lots of implicit information. 
An important direction of research here is how information is preserved as we increase the complexity of the prompt for specific tasks. 
For example, by asking a specific questions following a taxonomy, we could observe how much information is implicit in the model.
This in turn might inform how to design the prompt, and condition the LLM to maximise the quality of the desired output.
While different models and prompts greatly vary in capabilities and output, methodologies aimed to evaluate and analyse specific use-cases are valuable for designing special-purpose computational (creativity) systems.

Currently, we need to rely too heavily on chance to produce relevant and creative output.
These recommendations should steer us towards more informed and useful creative behaviours of LLMs.

\section{Related Work}
While this work does not focus on building a creative system, here we briefly outline several analogies and insights regarding the overlap between LLMs and narrative generation programs.
Programs that generate natural language and narratives have been an active research area.
Early examples of narrative generation programs are TALE-SPIN \cite{Meehan1977}, TAILOR \cite{SmithWitten1991}, and MINSTREL \cite{Turner1994}.
These programs are in some sense limited, since they rely on predetermined templates that are filled according to various rules and associations in a knowledge base. 
However, some stories generated by TALE-SPIN are surprisingly similar to what a LLM might generate for simple prompts (see \nameref{supplementary-materials}).
Another example that draws a parallel with LLMs is The Grandmother program \cite{Casebourne1996} which uses an interactive process for the user to provide additional knowledge, this is similarly to the idea of in-context learning \cite{Brown2020}.
The limitations of the symbolic approaches and reliance on chance to generate interesting stories are partly because the programs are unable to observe and evaluate their outputs.
MEXICA \cite{PerezyPerezSharples2001} is a narrative generation program that aims to tackle this with an engagement-reflection loop.
This can be mimicked in LLMs by asking to improve the previous output and giving hints, but this has limited success, as we do not know how or why the LLM improves a story.

Markov chains are a technique applied to narrative generation that is more akin to LLMs.  
While simple low-order Markov models generally perform poorly, variants have been applied successfully \cite{HarrisonPurdyRiedl2021,vanStegerenTheune2019,Ammanabrolu2019}.
LLMs, however, have largely displaced these approaches and easily appear to outperform on most dimensions.
There are plenty of examples \cite{Yuan2022,Kelly2023,MendezGervas2023,Ye2023} that introduce certain procedures complementing the LLM to perform the task. 
Perhaps, it is more appropriate to view the LLM as a ``mere'' component or stage in a larger creative system.

\section{Conclusion}
We present a two-fold empirical study to investigate how temperature affects the creativity of LLM-generated stories. 
Our methodology leverages the idea of the exemplar, inspired by prototype and exemplar theory, as a point of departure for evaluation.
In summary, our findings indicate that:
\begin{itemize}
    \item When generating a limited number of samples, temperature increases the chance to generate more variety, but overall, does not enable access to a larger slice of the probability distribution.
    \item We only observe a weak positive correlation between novelty and temperature, and a moderately negative correlation between coherence and temperature, implying a novelty and coherence trade-off.
\end{itemize}
The influence of temperature is far more nuanced and weak than ``the creativity parameter'' claim suggests.
To enable the potential of LLM creativity, we highlight the following research directions:
\begin{itemize}
    \item Explore what a benchmark for LLM creativity for evaluation at scale should be.
    \item Design advanced decoding strategies specifically for creative purposes.
    \item Methods to probe the implicit information captured by LLMs to understand and assess the effect of prompts and guide prompt design.
\end{itemize}

\section{Supplementary Materials}\label{supplementary-materials}
The appendix, code, statistical analysis, data, and generated stories can be found at \url{https://github.com/maxpeeperkorn/creativity-parameter}.

\section{Acknowledgements}
The study was approved by the Central Research Ethics Advisory Group (CREAG) of the University of Kent.
The work of MP is supported by a University of Kent Graduate Teaching Assistantship and a Cultuurfonds Award.
The work of DB is supported by a Discovery Grant from the Natural Sciences and Engineering Research Council of Canada.  

\bibliographystyle{iccc}
\bibliography{iccc}

\end{document}